%% file: main.tex
\pdfoutput=1  
\documentclass{article}

\usepackage[nonatbib,final]{neurips_2021}

\usepackage[utf8]{inputenc} 
\usepackage[T1]{fontenc}    
\usepackage{url}            
\usepackage{booktabs}       
\usepackage{amsfonts}       
\usepackage{nicefrac}       
\usepackage{microtype}      
\usepackage{xcolor}         
\usepackage{graphicx}
\usepackage{xspace}
\usepackage{comment}
\usepackage{multirow}
\usepackage{amsmath}
\usepackage{mathtools}
\usepackage{wrapfig}
\usepackage{bm}  
\usepackage[skip=5pt]{caption}
\usepackage[pagebackref,breaklinks,colorlinks]{hyperref}
\usepackage[capitalise]{cleveref}
\usepackage{etoc}
\usepackage[title]{appendix}
\usepackage{longtable}

\hypersetup{
    colorlinks=true,
    linkcolor=blue,
    filecolor=magenta,      
    urlcolor=cyan,
}

\newcommand*{\ourmodel}{\emph{NeRF$--$}\@\xspace}
\newcommand*{\baseline}{\emph{baseline NeRF}\@\xspace}

\newcommand*{\eg}{\emph{e.g.,}\@\xspace}
\newcommand*{\ie}{\emph{i.e.,}\@\xspace}

\definecolor{colmap_traj_colour}{HTML}{BF00BF}  

\usepackage[ruled]{algorithm2e} 

\SetAlFnt{\small}
\SetAlCapFnt{\small}
\SetAlCapNameFnt{\small}
\SetAlCapHSkip{0pt}

\newcommand{\myparagraph}[1]{\textbf{#1 \hspace{1ex}}}

\title{NeRF$--$: Neural Radiance Fields Without \\Known Camera Parameters}

\author{%
  Zirui Wang \\
  Active Vision Lab\\
  University of Oxford\\
  \texttt{ryan@robots.ox.ac.uk} \\
  \And
  Shangzhe Wu \\
  Visual Geometry Group \\
  University of Oxford \\
  \texttt{szwu@robots.ox.ac.uk} \\
  \AND
  Weidi Xie \\
  Visual Geometry Group \\
  University of Oxford \\
  \texttt{weidi@robots.ox.ac.uk} \\
  \And
  Min Chen \\
  e-Research Centre \\
  University of Oxford \\
  \texttt{min.chen@oerc.ox.ac.uk} \\
  \And
  Victor Adrian Prisacariu \\
  Active Vision Lab \\
  University of Oxford \\
  \texttt{victor@robots.ox.ac.uk} \\
}

\begin{document}

\maketitle
\input{0_abstract}

\input{1_intro}
\input{2_related}
\input{3_preliminary}
\input{4_method}
\input{5_dataset}
\input{6_experiments}

\input{7_conclusions}

\section*{Acknowledgement}
Shangzhe Wu is supported by Facebook Research.
Weidi Xie is supported by Visual AI (EP/T028572/1).
The authors would also like to thank Tim Yuqing Tang for insightful discussions and proofreading.

\appendix
\begin{appendices}
\input{9_supp}
\end{appendices}

\begin{small}
\bibliographystyle{unsrt}
\bibliography{bibliography}
\end{small}

\end{document}


\maketitle

\appendix
\input{supmat/01}


\begin{small}
\bibliographystyle{unsrt}
\bibliography{bibliography}
\end{small}

%% file: 0_abstract.tex
\begin{abstract}
Considering the problem of novel view synthesis~(NVS) from only a set of 2D images, we simplify the training process of Neural Radiance Field~(NeRF) on forward-facing scenes by removing the requirement of known or pre-computed camera parameters, including both intrinsics and 6DoF poses.
To this end, we propose \textit{NeRF$--$}, with three contributions:
\emph{First}, we show that the camera parameters can be jointly optimised as learnable parameters with NeRF training, through a photometric reconstruction;
\emph{Second}, to benchmark the camera parameter estimation and the quality of novel view renderings, 
we introduce a new dataset of path-traced synthetic scenes, termed as Blender Forward-Facing Dataset (BLEFF);
\emph{Third}, we conduct extensive analyses to understand the training behaviours under various camera motions,
and show that in most scenarios, the joint optimisation pipeline can recover accurate camera parameters and achieve comparable novel view synthesis quality as those trained with COLMAP pre-computed camera parameters. Our code and data is available at \url{https://nerfmm.active.vision}.
\end{abstract}

%% file: 1_intro.tex
\section{Introduction}
Generating photo-realistic images from arbitrary viewpoints not only requires to understand the 3D scene geometry but also complex viewpoint-dependent appearance resulted from sophisticated light transport phenomena.
One way to achieve this is by constructing a 5D plenoptic function that directly models the light passing through each point in space~\cite{Adelson91theplenoptic}~(or a 4D light field~\cite{Gortler1996Lumigraph,Levoy1996light} if we restrict ourselves outside the convex hull of the objects of interest).
Unfortunately, physically measuring such a plenoptic function is usually not feasible in practice. 
As an alternative, Novel View Synthesis~(NVS) aims to generate the unseen views from a small set of images captured from diverse viewpoints.

Recently, Neural Radiance Fields (NeRF)~\cite{mildenhall2020nerf, barron2021mip, martinbrualla2020nerfw, li2020neural} and Multi-Plane Images (MPI)~\cite{zhou2018stereo, tucker2020single, mildenhall2019llff, choi2019extreme, flynn2019deepview, wizadwongsa2021NeX} have demonstrated that, 
by optimising a volumetric scene representation on a sparse set of images, 
one can re-render the scene from novel viewpoints with impressive visual quality, 
including sophisticated view-dependent effects, such as specularity and transparency.

One limitation for these methods is the requirement of the camera parameters of the images at the training time, 
which is rarely accessible in real scenarios, \eg images taken with a mobile phone.
In practice, these methods usually pre-compute the camera parameters via conventional techniques. For example, NeRF and its variants~\cite{mildenhall2020nerf, barron2021mip, martinbrualla2020nerfw, li2020neural} adopt COLMAP~\cite{schonberger2016colmap} to estimate the camera parameters~(both intrinsics and extrinsics) for each input image.
This pre-processing step, apart from introducing additional complexity, 
can be potentially erroneous or simply fail,  
due to homogeneous regions, fast-changing view-dependent appearance, or degenerate solutions in linear equations from highly ambiguous camera trajectories \cite{luong1994stability, torr1998robust, hartley2003multiple}.

Our goal in this paper is to investigate novel view synthesis for forward-facing scenes with unknown camera parameters.
We introduce \ourmodel, a framework that treats camera poses and intrinsics as learnable parameters,
and jointly optimises them with a 3D scene representation~(as shown in Fig.~\ref{fig:teaser}).
Given only a sparse set of images captured in a forward-facing setup, our system can be trained 
1) to estimate camera parameters for each image and 
2) to learn a volumetric scene representation through a photometric reconstruction loss, 
achieving high fidelity NVS results comparable to existing two-stage COLMAP-NeRF pipelines
while removing the pre-processing step for camera parameters.

In order to establish a comprehensive evaluation of the method, we make two additional contributions:
1) we introduce a high-quality path-traced synthetic dataset, named Blender Forward Facing (BLEFF), to benchmark camera parameter estimation and novel view synthesis quality, which is publicly accessible for future research;
2) we conduct thorough analyses on our method and COLMAP-NeRF under various camera motions, 
showing that both systems can tolerate up to $\pm20^\circ$ of rotation and $\pm20\%$ translation perturbations, 
and that the joint optimisation is more favourable than COLMAP in translational perturbations, but less competitive in rotational perturbations.

%% file: 2_related.tex
\begin{figure}
    \centering
    \includegraphics[width=0.96\columnwidth]{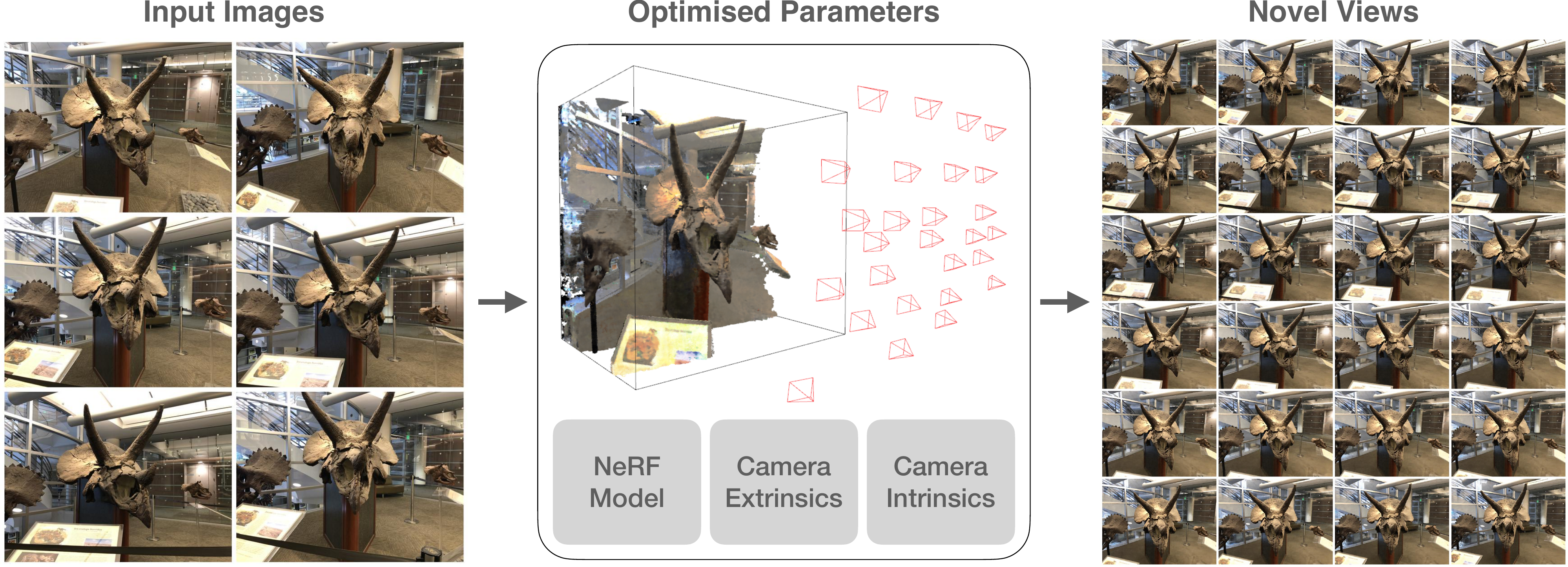}
    \caption{We propose \ourmodel, a NeRF-based framework for novel view synthesis without pose supervision.
    Specifically, our method jointly optimises camera poses and intrinsics of a set of forward-facing input images while training a NeRF model. \vspace{-15pt}
    }
    \label{fig:teaser}
\end{figure}

\section{Related Work}
We divide the related work roughly into two categories, 
one assuming camera parameters are input to NVS representations, 
and the other jointly estimating camera parameters with an NVS representation from uncalibrated images.

\myparagraph{With Known Camera Parameters} 
Images and their camera parameters are usually required by many novel view synthesis systems. 
In this line of work, classical methods Light Field Rendering~\cite{Levoy1996light} and Lumigraph~\cite{Gortler1996Lumigraph} model a simplified plenoptic function. 
\cite{Zhou2013mvps} integrates sophisticated hand-crafted material BRDF models into Signed Distance Functions~(SDF). FVS\cite{riegler2020free} and SVS\cite{riegler2020stable} combine meshes and deep features from images. 
Recently, dense volumetric representations have been proposed to enable smooth gradients for photometry-based optimisation and has shown to be promising for photo-realistic novel view synthesis of highly complex shapes and view-dependent appearance. These representations include Soft3D\cite{Soft3DReconstruction}, Multi-Plane Images~(MPI)\cite{zhou2018stereo, tucker2020single, mildenhall2019llff, choi2019extreme, flynn2019deepview, wizadwongsa2021NeX}, Scene Representation Networks~(SRN)\cite{sitzmann2019srns}, implicit surface \cite{Mescheder2019occnet,yariv2020multiview}, and Neural Radiance Fields~(NeRF)\cite{mildenhall2020nerf, martinbrualla2020nerfw, li2020neural, barron2021mip}. 
Despite these approaches differ in the way of scene representations, 
they all require input images to be associated with known camera parameters.


\myparagraph{With Unknown Camera Parameters}
The second line of work, mostly consisting of SLAM and SfM systems, aims to reconstruct the scene directly from RGB images by jointly estimating camera parameters and 3D geometries. 
For example, MonoSLAM\cite{davison2007monoslam} and ORB-SLAM~\cite{mur2015orb} 
reconstruct a point cloud map and estimate camera poses in real-time by associating feature correspondences, 
Similarly, DTAM~\cite{newcombe2011dtam}, SVO\cite{forster2014svo}, 
and LSD-SLAM~\cite{engel2014lsd} approach the SLAM task by minimising a photometric loss. 
SfM systems Bundler\cite{snavely2006bundler} and COLMAP \cite{schonberger2016colmap} can reconstruct a global consistent map and estimate camera parameters for large image sets, but sensitive to initial image pairs.
Although these methods only require RGB images as input, they often assume diffuse surface appearance in order to establish correspondences and cannot recover view-dependent appearance, hence resulting in unrealistic novel view rendering.



Concurrent to our work, several methods~\cite{Lin2020inerf, meng2021gnerf, lin2021barf} 
have developed similar ideas on estimating camera parameters with a NeRF model. 
These methods either requires intrinsics as input~\cite{meng2021gnerf, lin2021barf}, 
or assumes a pre-trained NeRF is available~\cite{Lin2020inerf}, 
In contrast, 
we propose to jointly optimise camera poses, intrinsics and a NeRF in an end-to-end manner on forward-facing scenes. As a result, we show the proposed framework is able to produce photo-realistic view synthesis that are comparable to two-stage NVS systems.



%% file: 3_preliminary.tex
\begin{figure*}[t]
\centering
\includegraphics[width=0.9\textwidth]{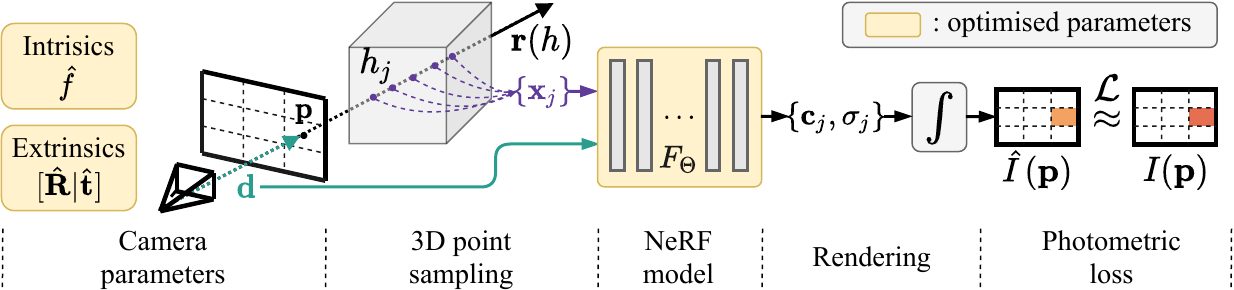}
\caption{Our pipeline jointly optimises a NeRF model and the camera parameters of the input images by minimising the photometric reconstruction errors.
To render a pixel $\mathbf{p}$ from NeRF, given the optimised camera parameters ($\hat{f}$, $\hat{\mathbf{R}}$, $\hat{\mathbf{t}}$), we feed 3D points $\mathbf{x}_j$ sampled along the camera ray together with the viewing direction $\mathbf{d}$ into NeRF $F_\Theta$, and aggregate the output radiance $\mathbf{c}_j$ and densities $\sigma_j$ to obtain its colour $\hat{I}(\mathbf{p)}$.
The entire pipeline can be trained end-to-end using only RGB images with unknown cameras as input.}
\label{fig:method_pipeline}
\vspace{-0.3cm}
\end{figure*}

\section{Preliminary}\label{sec:preliminary}
Given a set of images $\mathcal{I} = \{I_1, I_2, \dots, I_N\}$ captured from $N$ sparse viewpoints of a scene, 
with their associated camera parameters $\Pi = \{\pi_1, \pi_2, \dots, \pi_N\}$, including both intrinsics and 6DoF poses, the goal of novel view synthesis is to come up with a scene representation 
that enables the generation of realistic images from novel and unseen viewpoints.

In this paper, 
we follow the approach proposed in NeRF ~\cite{mildenhall2020nerf}.
Specifically, NeRF adopts an neural representation to construct a radiance field from sparse input views, where the view-dependent appearance is modelled by a continuous function $F_\Theta: (\mathbf{x}, \mathbf{d}) \rightarrow (\mathbf{c}, \sigma)$, which maps a 3D location $\mathbf{x} = (x, y, z)$ and a viewing direction $\mathbf{d}=(\theta, \phi)$ to a radiance colour $\mathbf{c}=(r, g, b)$ and a density $\sigma$. 

To render an image from the NeRF model, 
the colour at each pixel $\mathbf{p} = (u, v)$ on the image plane $\hat{I}_i$ is obtained by a rendering function $\mathcal{R}$, aggregating the radiance along a ray shooting from the camera position $\mathbf{o}_i$, 
passing through the pixel $\mathbf{p}$ into the volume~\cite{max1995optical, Gortler1996Lumigraph}:
    \begin{equation}
        \hat{I}_i(\mathbf{p}) = \mathcal{R}(\mathbf{p}, \pi_i | \Theta) 
        = \int_{h_n}^{h_f} T(h) \sigma\left(\mathbf{r}(h)\right) \mathbf{c}\left(\mathbf{r}(h), \mathbf{d}\right) \, dh,
    \label{eq:preliminary_render_pixel}
    \end{equation}
where 
$T(h) = \exp(-\int_{h_n}^{h}\sigma(\mathbf{r}(s))\, ds)$
denotes the accumulated transmittance along the ray,
\ie the probability of the ray travelling from $h_n$ to $h$ without hitting any other particle, 
and $\mathbf{r}(h) = \mathbf{o} + h\mathbf{d}$ denotes the camera ray that starts from camera origin $\mathbf{o}$ and passes through $\mathbf{p}$, controlled by the camera parameter $\pi_i$, with near and far bounds $h_n$ and $h_f$. 
In practice, the integral in~\cref{eq:preliminary_render_pixel} is approximated by accumulating radiance and densities of a set of sampled points along a ray.
With this implicit scene representation $F_\Theta(\mathbf{x},\mathbf{d})$ and a differentiable renderer $\mathcal{R}$, 
NeRF can be trained by minimising the photometric error $\mathcal{L} = \sum_i^N ||I_i - \hat{I}_i||^2_2$ between the observed views and synthesised ones $\hat{\mathcal{I}} = \{\hat{I}_1,...,\hat{I}_N\}$ under known camera parameters:
\begin{equation}
    \Theta^* = \arg\min_{\Theta} \mathcal{L}(\hat{\mathcal{I}} | \mathcal{I}, \Pi).
\end{equation}
To summarise, NeRF represents a 3D scene as a radiance field parameterised by MLPs, which is trained by minimising the discrepancy between the observed and rendered images. 
Note that, the camera parameters $\pi_i$ for these given images are required for training, 
which are usually estimated by SfM packages, such as COLMAP \cite{schonberger2016colmap}.
For more details of NeRF, we refer the readers to \cite{mildenhall2020nerf}.


%% file: 4_method.tex
\begin{figure}[t]
    \centering
    \includegraphics[width=0.99\textwidth]{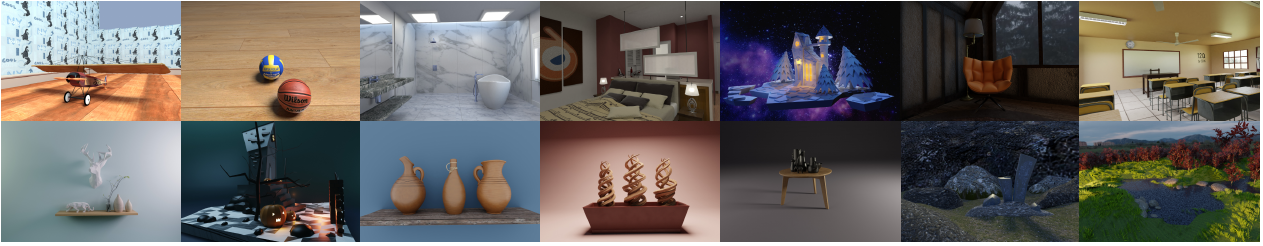}
    \caption{Thumbnails of BLEFF dataset.}
    \vspace{-0.3cm}
    \label{fig:dataset_thumbnails}
\end{figure}

\section{Method}
In this paper, 
we investigate the training framework for novel view synthesis without known camera parameters.
Our framework jointly optimises the camera parameters and a scene representation from a set of input RGB images $\mathcal{I}$. 
We make two assumptions: 
{\em First}, all images are captured in a forward-facing setup with certain amount of rotation and translation flexibility, as described in LLFF~\cite{mildenhall2019llff} and original NeRF\cite{mildenhall2020nerf}. 
{\em Second}, all images are captured with same intrinsic parameters.
Mathematically, our proposed framework \ourmodel can be formulated as:
\begin{align}
    \Theta^*, \Pi^* &= \arg\min_{\Theta, \Pi} \mathcal{L}(\hat{\mathcal{I}}, \hat{\Pi} | \mathcal{I}),
\label{eq:our_loss}
\end{align}
where $\Pi$ refers to the camera parameters, including both intrinsics and 6DoF poses.
In addition to simplifying the original two-stage approach, this joint optimisation approach enables globally consistent reconstruction results, similar to the bundle adjustment used in conventional SfM pipelines~\cite{Triggs2000bundle} and SLAM systems~\cite{davison2007monoslam,newcombe2011dtam, engel2014lsd}.

In the following sections, we first introduce the representations for the camera parameters and then describe the process of the joint optimisation.

\subsection{Camera Parameters}
\myparagraph{Camera Intrinsics} can be expressed as a focal length $f$ and principle points $c_x$ and $c_y$ for a pinhole camera model.
Without losing generality, we consider the centre of sensor as the camera principle points, 
\ie $c_x \approx W/2$ and $c_y \approx H/2$, 
where $H$ and $W$ denote the height and the width of the image, 
and all input images are taken by the same camera. 
Thus, estimating the camera intrinsics only refers to finding the focal length $f$ in this case.

\myparagraph{Camera Poses} can be expressed as a camera-to-world transformation matrix $\mathbf{T}_{wc} = [\mathbf{R}|\mathbf{t}]$ in SE(3), 
where $\mathbf{R} \in \text{SO(3)}$ and $\mathbf{t} \in \mathbb{R}^3$ denote the camera rotation and translation respectively.
Specifically, optimising the translation vector $\mathbf{t}$ can be as simple as setting it as trainable parameters as it is defined in the Euclidean space. Whereas for the camera rotation, which is in SO(3) space, we adopt the axis-angle representation: 
$\bm{\phi} \coloneqq \alpha\mathbf{\omega}$, $\bm{\phi} \in \mathbb{R}^3$, where $\mathbf{\omega}$ and $\alpha$ denote a normalised rotation axis and a rotation angle $\alpha$ respectively.
A rotation matrix $\mathbf{R}$ can be recovered from the Rodrigues’ formula 
$\mathbf{R} = \mathbf{I} 
            + \frac{\sin(\alpha)}{\alpha}\bm{\phi}^{\wedge} 
            + \frac{1-\cos(\alpha)}{\alpha^2}(\bm{\phi}^{\wedge})^2$, 
where $(\cdot)^{\wedge}$ is the skew operator that converts a vector $\bm{\phi}$ to a skew matrix.
Till here, we can optimise the camera poses for each input image $I_i$, with trainable parameters $\bm{\phi}_i$ and $\mathbf{t}_i$.

\subsection{Joint Optimisation of Camera Parameters and Scene Representation}
Now we introduce the details for training \ourmodel, which jointly optimises camera parameters along with a NeRF model on a sparse set of forward-facing views.

Specifically, to render a pixel $p$ on a given image $I_i$, 
we shoot a ray $\hat{\mathbf{r}}_{i,p}(h) = \hat{\mathbf{o}}_i + h \hat{\mathbf{d}}_{i,p},$ 
from the camera position through the pixel $p$ into the radiance field, 
with the current estimates of the camera parameters $\hat{\pi}_i = (\hat{f}, \hat{\bm{\phi}}_i, \hat{\mathbf{t}}_i$), 
and the ray direction 
\begin{align}
    \hat{\mathbf{d}}_{i,p} = \hat{\mathbf{R}}_i
            \begin{pmatrix}
                (u - W/2) / \hat{f}\\
                -(v - H/2) / \hat{f}\\
                -1
            \end{pmatrix},
\label{eq:build_ray}
\end{align}
where $\hat{\mathbf{o}}_i = \hat{\mathbf{t}}_i$ and $\hat{\mathbf{R}}_i$ is computed from $\hat{\bm{\phi}}_i$ using the Rodrigues’ formula. 
Along the ray, we sample a number of 3D points $\{\mathbf{x}_j\}$ and evaluate the radiance colours $\{\mathbf{c}_j\}$ and the density values $\{\mathbf{\sigma}_j\}$ of these points from the NeRF model~$F_\Theta$.
To get the colour for that pixel, 
the rendering function \cref{eq:preliminary_render_pixel} is applied to aggregate the predicted radiance and densities along the ray.

During training, we randomly render $M$ pixels for each input image, and minimise the reconstruction loss between the rendered colours against the ground-truth colours on these pixels.
Note that, the entire pipeline is fully differentiable, 
which enables the jointly optimisation of NeRF and the camera parameters $\{\pi_i\}$ by minimising the reconstruction loss.



%% file: 5_dataset.tex
\section{Blender Forward Facing Dataset}
    \begin{wrapfigure}{r}{0.4\columnwidth}
        \vspace{-0.4cm}
        \includegraphics[width=1.0\linewidth]{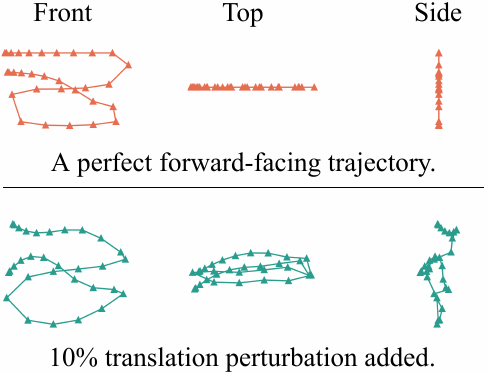}
        \caption{Dataset trajectory generation.}
        \label{fig:dataset_trajs}
        \vspace{-0.5cm}
    \end{wrapfigure}

Evaluating the performance of a joint optimisation system like ours requires both images and their ground truth camera parameters, which are not easy to acquire for real captured images. 
To that end, 
we introduce a synthetic dataset, Blender Forward Facing (BLEFF), 
containing 14 path-traced scenes\footnote{We modified open sourced blender files downloaded from \url{https://www.blendswap.com/}. 
Most blender files come with CC-BY-3.0 licence. Details see our supplementary.}, with each rendered in multiple levels of rotation and translation perturbations.
We will make this dataset publicly available, expecting to foster future research on the joint optimisation of camera parameters and scene representations.


We now detail how we create the dataset. 
Starting with a perfect forward-facing camera trajectory (top row in \cref{fig:dataset_trajs}), \ie all cameras looking forward and moving in $xy$-plane ($z$=0), 
we generate 16 trajectory variants for each scene by adding 5/5/6 levels of rotation/translation/full-6DoF 
perturbations on the perfect trajectory. 
We denote these trajectory variants via a $t_{aaa}r_{bbb}$ notation, 
where $t_{aaa}$ and $r_{bbb}$ denote $\pm aaa\%$ of translation and $\pm bbb$ degree of rotation perturbed, respectively. Since each synthetic scene is in different scale, 
we add translation perturbation proportional to the maximum dimension of a trajectory, hence the $\%$ notation.
\cref{fig:dataset_thumbnails} shows image examples of BLEFF and \cref{fig:dataset_trajs} presents a $t_{010}$ trajectory example. 

\myparagraph{Background} 
It is surprisingly difficult to find suitable datasets to 
jointly evaluate the performance of camera parameters and a NVS model at the same time. 
In general, three types of dataset are available, but cannot meet our requirements. 
The first type focuses on NVS rendering quality with their camera parameters estimated from SfM and SLAM packages\cite{schonberger2016colmap, mur2015orb}. 
These datasets include LLFF-NeRF\cite{mildenhall2019llff}, RealEstate10K\cite{zhou2018stereo}, Spaces\cite{flynn2019deepview}, Shiny\cite{wizadwongsa2021NeX}, and Tanks \& Temples\cite{knapitsch2017tanks}. 
Without ground truth camera parameters, it is impossible to evaluate the camera pose estimation accurately. 
The second type is designed to benchmark SLAM systems, \eg~TUM-RGBD\cite{sturm2012benchmark} and ETH3D\cite{schops2019badslam}, 
with ``ground truth'' camera parameters being estimated from an external motion capture system. 
However, we find the time lags between motion capture system and the captured images are unignorable in NVS tasks
and often lead to blurry renderings. 
The third type, \eg~SceneNet RGB-D\cite{mccormac2016scenenet} and InteriorNet\cite{li2018interiornet}, 
offers large-scale simulated indoor images but lacks of scene diversities.

%% file: 6_experiments.tex
\section{Experiments}

\subsection{Setups}\label{sec:exp_setup}
\myparagraph{Dataset}
Including our BLEFF dataset, we evaluate \ourmodel on three datasets:
\textbf{LLFF}: We first conduct experiments on the same forward-facing dataset as that in NeRF, namely, LLFF-NeRF~\cite{mildenhall2019llff}, which has 8 forward-facing scenes captured by mobile phones or consumer cameras, each containing 20-62 images. In all experiments, we follow the official pre-processing procedures and train/test splits, \ie~the resolution of the training images is $756 \times 1008$, and every $8$-th image is used as a test image.
\textbf{RealEstate10K}: To understand the behaviour of NVS in real-life capture scenarios, we additionally collected a number of diverse scenes extracted from the short video segments in RealEstate10K~\cite{zhou2018stereo}. In particular, the image resolution in these sequences is $1080 \times 1920$ and the frame rate is 30 fps. We sub-sample the frames and reduce the frame rates to 5 fps, with each sequence containing 14-43 images.
\textbf{BLEFF}: We use BLEFF to evaluate camera parameter estimation accuracy and NVS rendering quality, as well as to explore the breaking point of the \ourmodel and \baseline. All scenes in BLEFF come with 31 images in $1040 \times 1560$ resolution, with 27 for training and 4 for testing, and focal lengths set to 1040 pixels in both horizontal and vertical directions. In the experiments below, all BLEFF images are downsized by half, \ie images are in $520 \times 780$ and the ground truth focal length $f_{gt}$ is 520 for training and testing.

\myparagraph{Metrics}
\label{sec:metrics}
We evaluate the proposed framework from two aspects:
First, to measure the quality of novel view rendering, 
we adopt common metrics: Peak Signal-to-Noise Ratio (PSNR), Structural Similarity Index Measure(SSIM)~\cite{wang2004image} and Learned Perceptual Image Patch Similarity (LPIPS)~\cite{zhang2018perceptual};
Second, we evaluate the accuracy of the optimised camera parameters, including focal length, rotation and translation. For focal length evaluation, 
we report the absolute error in the metric of pixels.
For the camera poses, we follow the evaluation protocol of Absolute Trajectory Error (ATE)~\cite{Zhang18ATE_ethz,sturm2012benchmark}, 
aligning two sets of pose trajectories globally with a similarity transformation Sim(3) 
and reporting the rotation angle between two rotations and the absolute distance between two translation vectors.

\myparagraph{Implementation Details}
We implement our framework in PyTorch following the same architecture as original NeRF, except that, for computation efficiency, we:
(a) do not use the hierarchical sampling strategy;
(b) reduce the hidden layer dimension from 256 to 128; and
(c) sample only 1024 pixels from every input image and 128 points along each ray.
We use Kaiming initialisation~\cite{he2015delving} for the NeRF model, and initialise all cameras to be at origin looking at $-z$ direction, 
with focal length $f$ to be the image width.
We use three separate Adam optimisers for NeRF, camera poses and focal lengths respectively, all with an initial learning rate of $0.001$. The learning rate of the NeRF model is decayed every $10$ epochs by multiplying with $0.9954$ (equivalent to a stair-cased exponential decay), and learning rates of the pose and focal length parameters are decayed every $100$ epochs with a multiplier of $0.9$. All models are trained for $10000$ epochs unless otherwise specified. More technical details are included in the supplementary material.

\begin{figure}[t]
    \centering
    \includegraphics[width=0.99\textwidth]{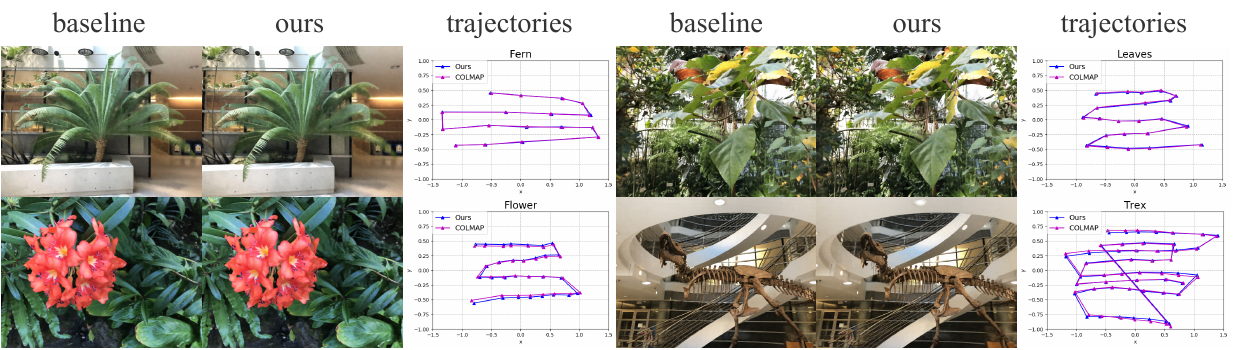}
    \caption{Qualitative comparison between our \ourmodel model with unknown cameras and the \baseline on LLFF-NeRF dataset.
    For each example, on the left, we show the synthesised novel views from the \baseline and from our model;
    on the right, we compare our optimised camera trajectories with the ones estimated from COLMAP, aligned with a Sim(3) transformation.
    Our proposed \ourmodel recovers similar camera poses with COLMAP and produces high quality novel views comparable to the \baseline. Full results can be found in the supplementary.
    }
    \label{fig:img_quality_nerf}
    \vspace{-0.6cm}
\end{figure}

\subsection{Results}


\subsubsection{On Novel View Synthesis Quality} \label{sec:nvs_result}
    \begin{wrapfigure}{r}{0.5\columnwidth}
        \vspace{-1.3cm}
        \centering
        \includegraphics[width=0.5\columnwidth]{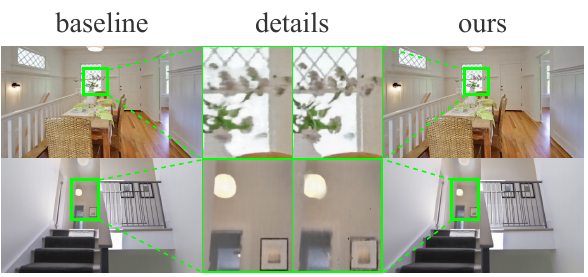}
        \caption{Qualitative comparison of NVS quality between our model and the baseline on RealEstate10K. Our model produces better results on the first examples but slightly worse results on the second example, indicating comparable performance overall compared to baseline.}
        \label{fig:img_quality_nerf_RE10K}
        \vspace{-0.8cm}
    \end{wrapfigure}
We compare the novel view rendering results from \baseline, where camera parameters are estimated from COLMAP, 
and our proposed model \ourmodel, which jointly optimises the camera parameters and the 3D scene representation.
We report the quantitative evaluations in~\cref{tab:baseline_vs_ours_LLFF} and visual results in~\cref{fig:img_quality_nerf} and \cref{fig:img_quality_nerf_RE10K}. 
Overall, our joint optimisation model, which does not require camera parameters as inputs, 
achieves similar NVS quality compared to the \baseline model
(perceptual quality metric $\Delta$SSIM and $\Delta$ LPIPS = 0.05).
This confirms that jointly optimising the camera parameters and 3D scene representation is indeed possible.
Nevertheless, we observe that for both the \textit{Orchids} and the \textit{Room}, 
our \ourmodel model produces slightly worse results compared to the \baseline. 
One reason can be our joint optimisation for these two scenes might have trapped into local minima, 
ending up with inaccurate intrinsics, as we notice from \cref{tab:baseline_vs_ours_LLFF} that the difference between optimised camera focal lengths and COLMAP estimation are most noticeable for these two scenes (199.3 and 331.8).

    \begin{table}[]
    \centering
    \resizebox{\textwidth}{!}{%
    \begin{tabular}{llcclcclcclccc}
    \hline
    \multirow{2}{*}{Scene} &  & \multicolumn{2}{c}{SSIM$\uparrow$} &  & \multicolumn{2}{c}{LPIPS$\downarrow$} &  & \multicolumn{2}{c}{PSNR$\uparrow$} &  & \multicolumn{3}{c}{Camera Parameters Difference} \\ \cline{3-4} \cline{6-7} \cline{9-10} \cline{12-14} 
     &  & colmap & ours &  & colmap & ours &  & colmap & ours &  & $\Delta$rot (deg) & $\Delta$tran & $\Delta$focal (pixel) \\ \hline
    Fern &  & 0.64 & 0.61 &  & 0.47 & 0.50 &  & 22.22 & 21.67 &  & 1.78 & 0.029 & 153.5 \\
    Flower &  & 0.71 & 0.71 &  & 0.36 & 0.37 &  & 25.25 & 25.34 &  & 4.84 & 0.016 & 13.2 \\
    Fortress &  & 0.73 & 0.63 &  & 0.38 & 0.49 &  & 27.60 & 26.20 &  & 1.36 & 0.025 & 144.1 \\
    Horns &  & 0.68 & 0.61 &  & 0.44 & 0.50 &  & 24.25 & 22.53 &  & 5.55 & 0.044 & 156.2 \\
    Leaves &  & 0.52 & 0.53 &  & 0.47 & 0.47 &  & 18.81 & 18.88 &  & 3.90 & 0.016 & 59.0 \\
    Orchids &  & 0.51 & 0.39 &  & 0.46 & 0.55 &  & 19.09 & 16.73 &  & 4.96 & 0.051 & 199.3 \\
    Room &  & 0.87 & 0.84 &  & 0.40 & 0.44 &  & 27.77 & 25.84 &  & 2.77 & 0.030 & 331.8 \\
    Trex &  & 0.74 & 0.72 &  & 0.41 & 0.44 &  & 23.19 & 22.67 &  & 4.67 & 0.036 & 89.3 \\ \hline
    Mean &  & 0.68 & 0.63 &  & 0.42 & 0.47 &  & 23.52 & 22.48 &  & 3.73 & 0.031 & 143.3 \\ \hline
    \end{tabular}%
    }
    \caption{Quantitative comparison between our model and the \baseline on LLFF-NeRF dataset. 
    One the left, we show the NVS quality of our method with unknown cameras, 
    achieving comparable results to \baseline: $\Delta$SSIM and $\Delta$LPIPS = 0.05, $\Delta$PSNR = 1.0. 
    Note that the \baseline numbers in this table is lower than the numbers reported in original paper, 
    as we employed a smaller NeRF model for both \baseline and \ourmodel for computation efficiency. 
    On the right, we report the difference between our optimised camera parameters and ones computed from COLMAP. 
    The result suggests our optimised camera poses are close to COLMAP estimations.}
    \label{tab:baseline_vs_ours_LLFF}
    \vspace{-0.5cm}
    \end{table}

\subsubsection{On Camera Parameter Estimation} \label{sec:camera_result}
Considering a forward-facing image capturing process in real life, 
where a user is likely to introduce unstable camera motions, 
to simulate this effect, we select $t_{010}r_{010}$ subset in BLEFF~\footnote{Suppose a trajectory of interest is bounded by a $1m\times1m$ square, $\pm10\%$ translation perturbation offers about $\pm10\% \times 1.414m \Rightarrow 28.3cm$ flexibility in $z$ direction during capturing.}.
\cref{tab:focal_ate_BLEFF} lists the L1 difference on the estimated focal length, 
and metrics on camera rotation and translation computed with the ATE toolbox~\cite{Zhang18ATE_ethz}. 
In summary, our rotation estimation error is about $5^\circ$, and focal length error is about 25 pixels, 
where as COLMAP has a slightly larger rotation error but lower focal length error. 
As a result, the NVS quality of our method is comparable with \baseline.

    \begin{table}[h]
    \centering
    \begin{tabular}{llccclcc}
    \hline
    \multirow{2}{*}{Method} &  & \multicolumn{3}{c}{Camera Parameters Error (vs. GT)} &  & \multicolumn{2}{c}{NVS Quality} \\ \cline{3-5} \cline{7-8} 
     &  & \multicolumn{1}{l}{$\Delta$focal (pixel)} & \multicolumn{1}{l}{$\Delta$rot (deg)} & $\Delta$tran &  & SSIM$\uparrow$ & \multicolumn{1}{l}{PSNR$\uparrow$} \\ \hline
    colmap &  & 14.89 & 13.65 & 0.0127 &  & 0.90 & 33.92 \\
    ours &  & 20.55 & 4.45 & 0.0654 &  & 0.90 & 33.24 \\ \hline
    \end{tabular}
    \caption{Quantitative evaluation of camera parameters estimation for both COLMAP and our method. We show that both the camera parameters estimation and NVS performance are comparable to COLMAP in BLEFF $t_{010}r_{010}$. A comprehensive list of results for each scene can be found in the supplementary.}
    \label{tab:focal_ate_BLEFF}
    \vspace{-0.4cm}
    \end{table}

    \begin{wrapfigure}{r}{0.5\columnwidth}
        \vspace{-0.4cm}
        \centering
        \includegraphics[width=0.5\columnwidth]{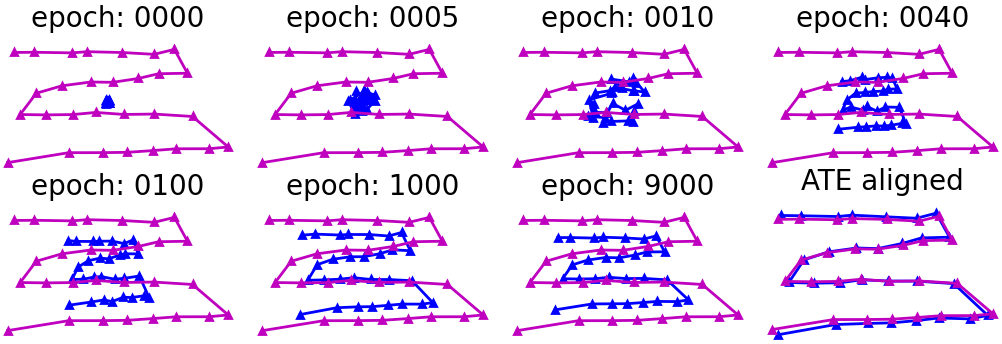}
        \vspace{-0.5cm}
        \caption{Pose optimisation during training, visualised on $xy$-plane
        (\textcolor{colmap_traj_colour}{purple: COLMAP}, \textcolor{blue}{blue: ours}).}
        \label{fig:pose_vis}
    \end{wrapfigure}
    
\myparagraph{Visualisation of Pose Optimisation}
For a better understanding of the optimisation process,
we visualize the camera poses at various training epochs for the scene \textit{Flower} from LLFF-NeRF dataset~(\cref{fig:pose_vis}).
Starting from identity, our optimised camera poses gradually converge towards COLMAP estimations after about 1000 epochs, subject to a similarity transformation between optimised camera parameters and those estimated from COLMAP.

\subsection{Discussion: Effects of Camera Motions} \label{sec:exp_analysis}
Considering \baseline and \ourmodel as two alternatives, this section demystifies when and why one method are more favourable than the other under various camera motion patterns. We expand this section from two aspects: First, we study the performance of our system and COLMAP under multiple levels of camera trajectory perturbations and explore the breaking point of both systems; Second, we examine system behaviours when a camera follows certain motion patterns during image capturing. As this section unfolds, we can see the necessity of discussing these common aspects in real-life capturing.

\subsubsection{Breaking Point in Forward-Facing Scenes} \label{sec:breakpt}
\begin{figure}[bh]
        \centering
        \includegraphics[width=1.0\columnwidth]{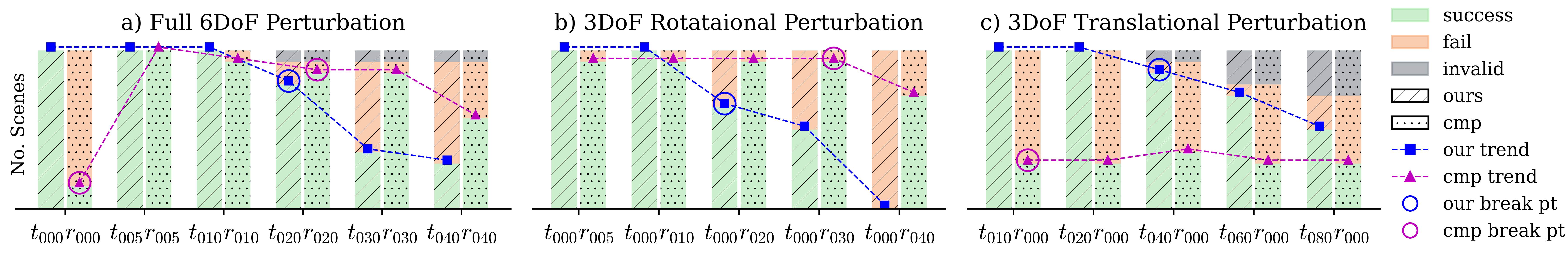}
        \caption{Breaking point analysis for camera parameter estimation in three groups of perturbation experiments: a) full 6DoF, b) 3DoF rotation, and c) 3DoF translation. COLMAP is shorten to "cmp" in the graph. We define a breaking point to the trajectory variant where a method starts failing in an experiment group. In group a) with full 6DoF noise, both methods start failing at $t_{020}r_{020}$. In 3DoF perturbations, our method performs more stable in translation perturbations but less stable in rotation perturbations. Note that COLMAP also faces degenerate issues with $t_{000}r_{000}$. Check \cref{sec:discuss_controlled_motion} for more details on degenerate cases.}
        \label{fig:discuss_break_point}
    \end{figure}

The forward-facing assumption we made, 
which in an ideal case, assumes all cameras looking to $-z$ direction and moving within $xy$-plane. 
However, this assumption is often over optimistic in real-life image capturing, 
it is thus interesting to study the robustness of both COLMAP and \ourmodel under multiple levels of trajectory perturbations (\cref{fig:discuss_break_point}).

\myparagraph{Setup} Three groups of perturbation experiments are conducted in this section (\cref{fig:discuss_break_point}): a) full 6DoF, b) 3DoF rotation, and c) 3DoF translation. We train all models for 3K epochs (not full 10K) and compare camera parameter estimations with ground truth with a criteria of mean rotation error $<20^\circ$ and focal length error $<50\%$ ground truth focal, as we observe severe NVS performance drop when the criteria is violated.

\myparagraph{Result}
With full 6DoF perturbation added, 
both \ourmodel and COLMAP start breaking when translation noise $>\pm20\%$ and rotation noise $>\pm20^\circ$. 
Our method is favourable than COLMAP in large translation perturbations, 
but underperforms COLMAP in large rotation cases. 
Two reasons account for this result. 
First, it is well known that feature descriptors can handle appearance changes better than photometric methods, 
hence the COLMAP's superior performance in large rotation perturbation. 
However, feature-based method can also be vulnerable to degenerate cases, 
resulting in the failure cases in translation perturbation experiments, 
as will be discussed in \cref{sec:discuss_controlled_motion}.

\subsubsection{Controlled Motion Patterns} 
\label{sec:discuss_controlled_motion}
    \begin{wrapfigure}{r}{0.4\columnwidth}
        \vspace{-0.4cm}
        \centering
        \includegraphics[width=1.0\linewidth]{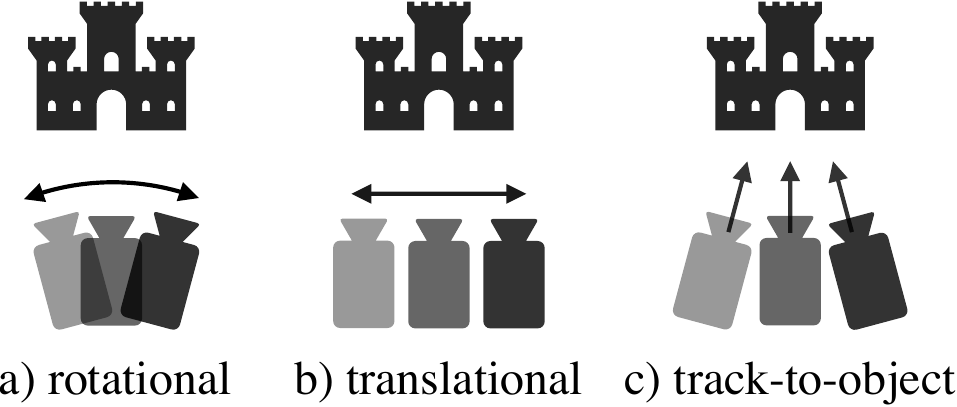}
        \caption{Three common camera motions in real-life capturing.}
        \label{fig:discuss_cam_motions}
        \vspace{-0.25cm}
    \end{wrapfigure}

In this section,
we study three controlled camera motion patterns (\cref{fig:discuss_cam_motions}): 
a) Rotational, where a camera motion contains mostly rotation with little translation. 
Note this is different from rotation perturbation experiment in \cref{sec:breakpt}, 
where the camera follows a planar trajectory but with perturbed viewing directions; 
b) Translational, in which a camera moves parallel to the camera sensor; 
c) Track-to-object, when a camera tracks an object. 
At the first glance, one may doubt if these controlled motions only exist in the synthetic world.
In fact, they can often happen in real-life image capturing, especially when a camera is attached to a gimbal, dolly, or a drone. 

In short, our method outperforms COLMAP in rotational and translation camera motions, but underperforms COLMAP when a camera tracks an object.

    \begin{wrapfigure}{r}{0.4\columnwidth}
        \vspace{-0.1cm}
        
        \begin{minipage}{0.4\columnwidth}
        \centering
        \includegraphics[width=1.0\linewidth]{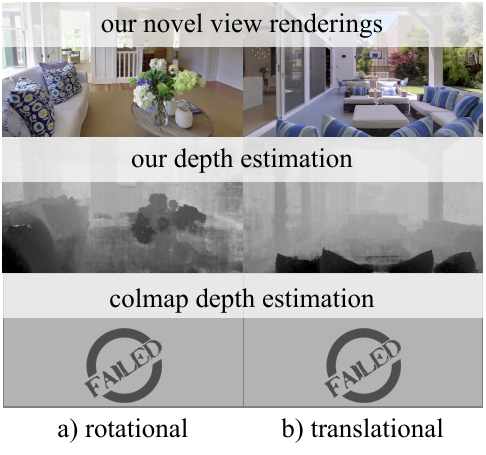}
        \caption{Rotational and translational scenes that are problematic to COLMAP in real image dataset RealEstate10K.}
        \label{fig:discuss_controlled_re}
        \end{minipage}
        \vspace{0.4cm}
        
        \begin{minipage}{0.4\columnwidth}
        \centering
        \includegraphics[width=1.0\linewidth]{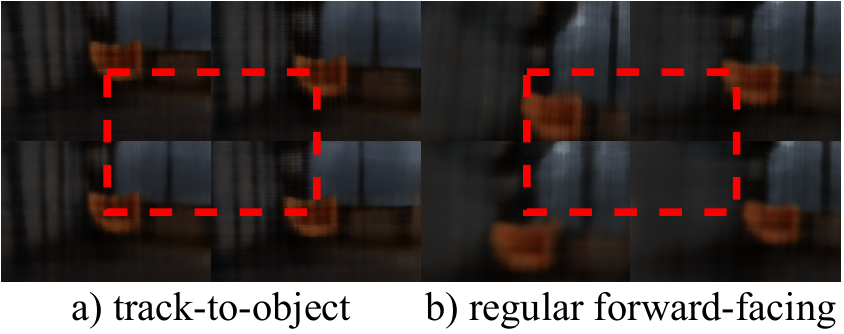}
        \caption{Rendered training views in two types of scenes at epoch 20. Unlike regular forward-facing scenes, where blurry rendered images are distinguishable even at early training stage, track-to-object scenes cannot provide enough photometric supervision from blurry rendered images, as the object position remains mostly unchanged (hinted by red dashed line) in all views.}
        \label{fig:discuss_track_obj}
        \end{minipage}
        \vspace{-1.3cm}
    \end{wrapfigure}
\myparagraph{Rotational and Translational} 
As shown in \cref{fig:discuss_break_point}c) and $t_{000}r_{000}$ column, 
COLMAP gives inaccurate camera parameter estimations in all translational experiments. 
Similar results are also observed in rotational motion. 
\cref{fig:discuss_controlled_re} presents two examples from the RealEstate10K dataset, 
where COLMAP fails to initialise or produces inaccurate estimations. 
For explanation, these results are known as degenerate cases while solving fundamental matrices from the point correspondences~\cite{luong1994stability, hartley2003multiple}. 
Moreover, in \cite{torr1998robust}, Torr et al. further point out that these degenerate cases are even more difficult to identify when matching outliers exist. 
Consequently, feature-matching-based SfM and SLAM methods tend to be vulnerable to these cases, 
whereas our photometry-based method handles them well. 

\myparagraph{Track-to-object} 
We observe an inaccurate joint optimisation from our model, 
when a camera is set to track to an object during capturing, 
even when the camera rotation is well below our breaking point ($\pm20^\circ$). 
\cref{fig:discuss_track_obj} illustrates such a situation. 
One hypothesis is, such track-to-object motion results in an effect that the object position and the image composition remain almost unchanged during capturing, which is not a problem for systems that work with high-res images directly, 
\eg COLMAP and SLAM systems. 
However, our method is supervised from the photometric loss between captured images and synthesised images. 
When the model is trained from scratch, synthesised images are blurry during most of the training time, 
producing little supervision for camera parameter optimisation. 

\subsection{Limitations and Future Work}
\label{sec:limitation}
    \begin{wrapfigure}{r}{0.18\columnwidth}
        \vspace{-0.4cm}
        \centering
        \includegraphics[width=1.0\linewidth]{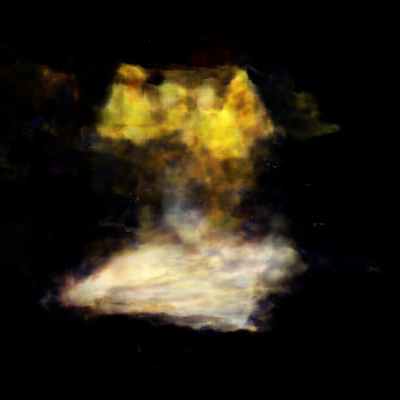}
        \caption{Failed \textit{Lego} $360^\circ$ scene.}
        \label{fig:failed_360}
        \vspace{-0.3cm}
    \end{wrapfigure}
Our system faces two limitations. 
\emph{Firstly}, as motioned in section \cref{sec:discuss_controlled_motion}, 
our method does not estimate meaningful camera parameters when images are captured in a track-to-object motion, 
as there is not enough supervision to guide the optimisation of camera parameters in such motions. 
\emph{Secondly}, we design our system to handle forward-facing scenes, 
which allows us initialising camera poses from identity matrices and optimising camera poses via gradient descent. 
However, this assumption restricts our pose estimation capability when the camera undergoes large rotations. 
As a result, camera motions with rotation perturbations larger than $\pm20^\circ$ might cause a failure (\cref{fig:discuss_break_point}), and our method certainly cannot handle $360^\circ$ scenes (\cref{fig:failed_360}).
As for future work, exploring relative pose estimation between image pairs and temporal relationships in sequences might enable the processing of such $360^\circ$ scenes.

%% file: 7_conclusions.tex
\section{Conclusions}

Our contribution in this work is threefold. 
\emph{First}, we present \ourmodel, 
a framework that jointly optimises camera parameters and scene representation for forward-facing scenes,
eliminating the need of pre-computing camera parameters, 
while achieving view synthesis results on par with the COLMAP-enabled NeRF baseline. 
\emph{Second}, we introduce the BLEFF datasets for evaluating such jointly optimised novel view synthesis system with ground-truth camera parameters and high-quality path-traced images.
\emph{Third}, we present extensive experimental results and demonstrate the viability of this joint optimisation framework under different camera trajectory patterns, even when COLMAP fails to estimate the camera parameters.
Despite its current limitations discussed above, 
our proposed joint optimisation pipeline has demonstrated promising results on this highly challenging task, 
and advanced one step towards end-to-end novel view synthesis on more general scenes.



%% file: 9_supp.tex
\section{Additional Results}
In this section, we provide additional results from 4 directions: 
1) NVS rendering quality;
2) Camera parameter estimation accuracy;
3) Full breaking point analysis results; and
4) Results of an optional refinement step.

\subsection{On NVS Quality}
\begin{figure}[h]
    \centering
    \includegraphics[width=\columnwidth]{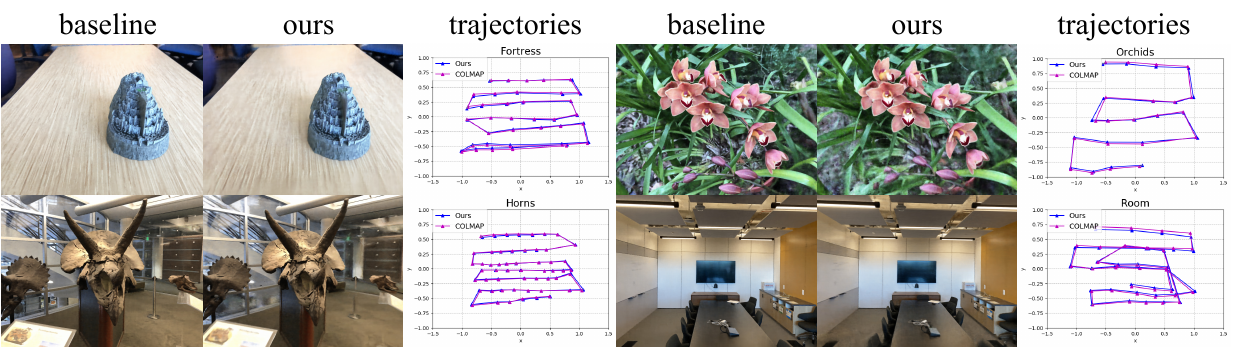}
    \caption{Additional Qualitatively Results on LLFF-NeRF Dataset. Our method achieves comparable NVS quality to the COLMAP-enabled NeRF while taking RGB images as the only input.}
    \label{fig:supp_llff_supp}
\end{figure}

\cref{fig:supp_llff_supp} presents additional qualitative results on LLFF-NeRF dataset. Our method achieves comparable novel view synthesis quality to the \baseline while taking RGB images as the only inputs. 

\vspace{-0.2cm}

\subsection{On Camera Parameter Estimation}
We report the full pose estimation results for Section 6.2.2. In short, our method produces accurate pose estimations and comparable NVS rendering quality to COLMAP-enabled NeRF ($\Delta\text{SSIM} = 0$).

\paragraph{Pose Ambiguity} 
We notice a special case \textit{Roundtable}, where there is a $179.10^\circ$ rotation error between COLMAP predicted poses and ground truth poses but the COLMAP-NeRF model still produces high quality NVS renderings. 
This is a case of ambiguous pose estimation as stated in \cite{oberkampf1996iterative,schweighofer2006robust}, where two sets of pose estimations are mirror images with respect to a plane. 
Nevertheless, as the estimated mirrored poses and the estimated 3D geometries from COLMAP are self-consistent, a NeRF model trained in this case can still provide plausible novel view synthesis, hence the high PSNR and SSIM in evaluation.

\begin{table}[bh]
\centering
\resizebox{\textwidth}{!}{%
\begin{tabular}{lcclcclcclcclcc}
\hline
\multirow{2}{*}{Scene} & \multicolumn{2}{c}{Focal Err. (pixel)} &  & \multicolumn{2}{l}{Rot. Err. (deg)} &  & \multicolumn{2}{c}{Tran. Err.} &  & \multicolumn{2}{c}{SSIM$\uparrow$} &  & \multicolumn{2}{c}{PSNR$\uparrow$} \\ \cline{2-3} \cline{5-6} \cline{8-9} \cline{11-12} \cline{14-15} 
 & colmap & ours &  & colmap & ours &  & colmap & ours &  & colmap & ours &  & colmap & ours \\ \hline
Airplane & 0.09 & 0.87 &  & 0.02 & 0.61 &  & 0.000 & 0.003 &  & 0.83 & 0.87 &  & 28.19 & 30.57 \\
Balls & 0.21 & 15.44 &  & 0.06 & 13.43 &  & 0.001 & 0.285 &  & 0.89 & 0.81 &  & 34.56 & 32.12 \\
Bathroom & 0.02 & 0.52 &  & 1.49 & 1.50 &  & 0.001 & 0.004 &  & 0.93 & 0.94 &  & 32.64 & 31.58 \\
Bed & 0.26 & 0.39 &  & 2.21 & 2.21 &  & 0.003 & 0.004 &  & 0.94 & 0.94 &  & 33.28 & 32.41 \\
Castle & 0.04 & 3.23 &  & 0.04 & 3.17 &  & 0.001 & 0.020 &  & 0.88 & 0.89 &  & 32.44 & 32.74 \\
Chair & 0.01 & 6.12 &  & 0.03 & 3.52 &  & 0.000 & 0.078 &  & 0.84 & 0.81 &  & 33.42 & 32.24 \\
Classroom & 0.15 & 2.29 &  & 0.35 & 8.14 &  & 0.002 & 0.032 &  & 0.90 & 0.86 &  & 27.49 & 25.14 \\
Deer & 1.93 & 6.29 &  & 0.42 & 6.17 &  & 0.015 & 0.166 &  & 0.99 & 0.99 &  & 41.44 & 42.01 \\
Halloween & 0.69 & 9.77 &  & 0.13 & 6.74 &  & 0.003 & 0.064 &  & 0.94 & 0.91 &  & 31.59 & 29.30 \\
Jugs & 0.05 & 0.16 &  & 0.22 & 2.30 &  & 0.004 & 0.065 &  & 0.98 & 0.99 &  & 40.76 & 42.53 \\
Root & 15.27 & 36.42 &  & 6.95 & 4.51 &  & 0.078 & 0.054 &  & 0.97 & 0.97 &  & 36.49 & 35.45 \\
Roundtable* & 189.64 & 206.10 &  & 179.10 & 9.68 &  & 0.069 & 0.139 &  & 0.99 & 0.99 &  & 45.13 & 39.88 \\
Stone & 0.04 & 0.11 &  & 0.04 & 0.12 &  & 0.000 & 0.001 &  & 0.79 & 0.86 &  & 29.95 & 31.74 \\
Valley & 0.02 & 0.05 &  & 0.04 & 0.25 &  & 0.000 & 0.001 &  & 0.73 & 0.75 &  & 27.53 & 27.66 \\ \hline
Mean & 14.89 & 20.55 &  & 13.65 & 4.45 &  & 0.013 & 0.065 &  & 0.90 & 0.90 &  & 33.92 & 33.24 \\ \hline
\end{tabular}%
}
\caption{Detailed quantitative camera parameter estimation results for Section 6.2.2. Our method produces accurate camera parameter estimations and comparable NVS quality to COLMAP-NeRF ($\Delta\text{SSIM} = 0$). \textit{Roundtable}* is a special case where COLMAP estimated poses are mirrored with the ground truth.
We tried 10 random seeds for COLMAP on this scene, 9/10 trials produce $179.1^\circ$ rotation error and 1/10 trials produce $0.71^\circ$ rotation error. Hence we report the $179.1^\circ$ here. More details see the text above.
}
\end{table}

\subsection{Breaking Point Analysis}
\begin{figure}[h]
    \centering
    \includegraphics[width=\columnwidth]{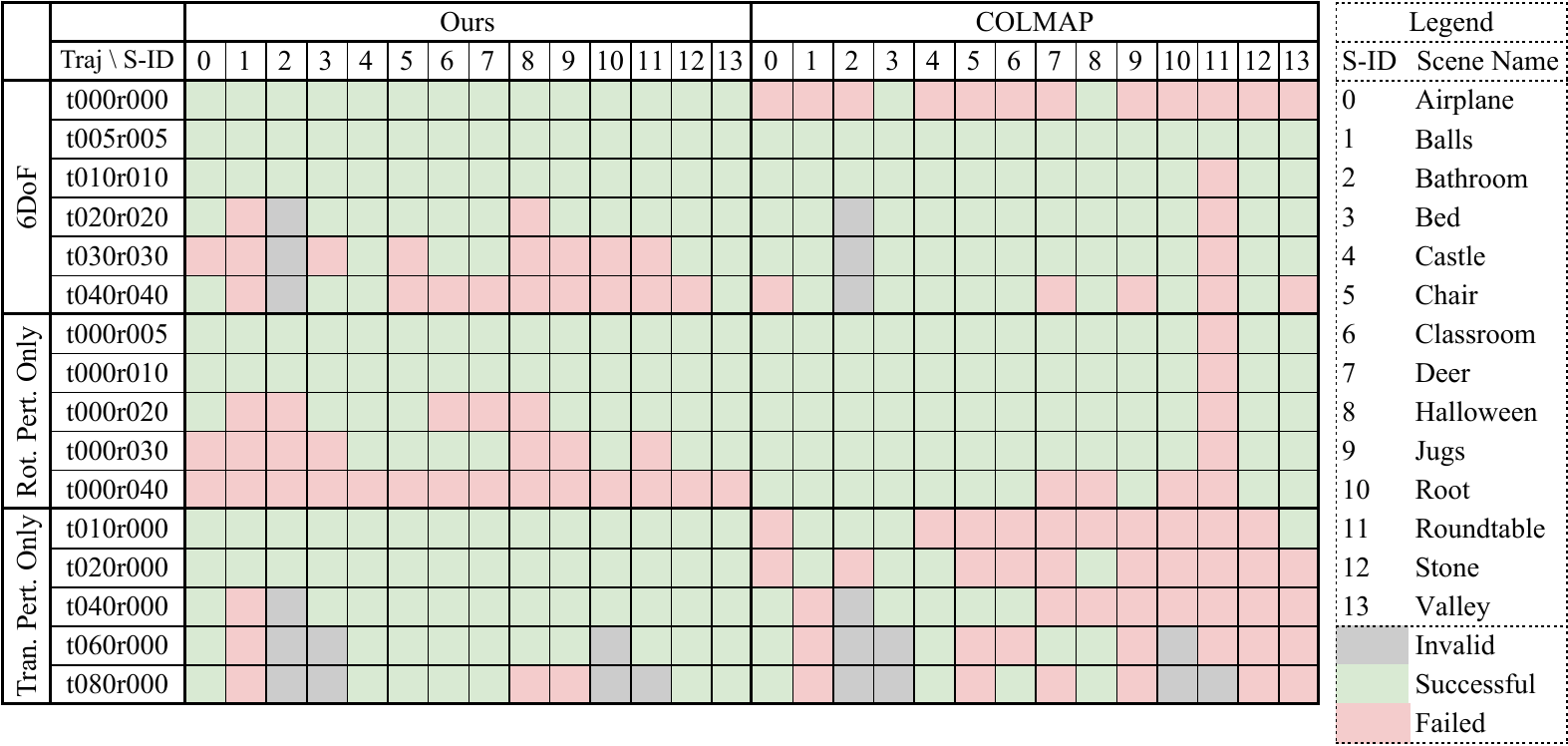}
    \caption{Per-scene camera parameter estimation results for breaking point analysis. Left: ours. Right: COLMAP. For each row (trajectory variant), the larger green area is, the better performance a method achieves in that trajectory. \textit{Invalid} (grey) denotes that a part of a camera trajectory is out of a blender scene after perturbed, resulting in meaningless image renderings, from which it is impossible for any method to recover camera parameters or scene geometries. We set a criteria of $\pm20^\circ$ rotation error and $\pm 0.5 f_{gt}$ focal length error to separate \textit{successful} (green) and \textit{failed} (red) camera parameter estimations.}
    \label{fig:supp_breakpt_full_table}
\end{figure}

In \cref{fig:supp_breakpt_full_table} we present a per-scene breakdown for the breaking point analysis in Section 6.3.1. Our method performs better than COLMAP in translation perturbation but worse in rotation perturbation.

\subsection{Optional Refinement}
We propose an optional refinement step to further improve the quality of the synthesised images.
Specifically, after the first training process is completed, we drop the trained NeRF model and re-initialise it with random parameters while keeping the pre-trained camera parameters.
We then repeat the joint optimisation using the pre-trained camera parameters as initialisation.
We find this additional refinement step generally leads to sharper images and improves the synthesis results, as evidenced by the comparison in~\cref{tab:supp_refine_llff}.

Additionally, the camera parameters can also be initialised with estimated values from external toolboxes, where they are available, and jointly refined during the training of the NeRF model.
We conduct experiments to refine the camera parameters estimated using COLMAP during NeRF training, and find the novel view results slightly improved through the joint refinement, as shown in~\cref{tab:supp_refine_llff}.

\begin{table}[hb]
\centering
\resizebox{\textwidth}{!}{%
\begin{tabular}{llcccclcccclcccc}
\hline
\multirow{2}{*}{Scene} &  & \multicolumn{4}{c}{SSIM$\uparrow$} &  & \multicolumn{4}{c}{LPIPS$\downarrow$} &  & \multicolumn{4}{c}{PSNR$\uparrow$} \\ \cline{3-6} \cline{8-11} \cline{13-16} 
 &  & colmap & ours & \multicolumn{1}{l}{colmap + r} & \multicolumn{1}{l}{ours + r} &  & colmap & ours & \multicolumn{1}{l}{colmap + r} & \multicolumn{1}{l}{ours + r} &  & colmap & ours & \multicolumn{1}{l}{colmap + r} & \multicolumn{1}{l}{ours + r} \\ \hline
Fern &  & 0.64 & 0.61 & 0.64 & 0.63 &  & 0.47 & 0.50 & 0.47 & 0.48 &  & 22.22 & 21.67 & 22.28 & 21.93 \\
Flower &  & 0.71 & 0.71 & 0.72 & 0.72 &  & 0.36 & 0.37 & 0.36 & 0.35 &  & 25.25 & 25.34 & 25.53 & 25.58 \\
Fortress &  & 0.73 & 0.63 & 0.72 & 0.71 &  & 0.38 & 0.49 & 0.38 & 0.38 &  & 27.60 & 26.20 & 27.13 & 27.37 \\
Horns &  & 0.68 & 0.61 & 0.68 & 0.64 &  & 0.44 & 0.50 & 0.44 & 0.48 &  & 24.25 & 22.53 & 24.44 & 23.18 \\
Leaves &  & 0.52 & 0.53 & 0.53 & 0.53 &  & 0.47 & 0.47 & 0.47 & 0.46 &  & 18.81 & 18.88 & 18.94 & 18.96 \\
Orchids &  & 0.51 & 0.39 & 0.52 & 0.41 &  & 0.46 & 0.55 & 0.45 & 0.53 &  & 19.09 & 16.73 & 19.31 & 17.03 \\
Room &  & 0.87 & 0.84 & 0.87 & 0.85 &  & 0.40 & 0.44 & 0.39 & 0.42 &  & 27.77 & 25.84 & 27.96 & 26.28 \\
Trex &  & 0.74 & 0.72 & 0.74 & 0.73 &  & 0.41 & 0.44 & 0.41 & 0.43 &  & 23.19 & 22.67 & 23.13 & 22.95 \\ \hline
Mean &  & 0.68 & 0.63 & 0.68 & 0.65 &  & 0.42 & 0.47 & 0.42 & 0.44 &  & 23.52 & 22.48 & 23.59 & 22.91 \\ \hline
\end{tabular}%
}
\caption{An optional camera parameter refinement (denote by '+r') can further improve NVS quality. During the refinement, camera parameters are first initialised from COLMAP estimations or from previous joint optimisations, and jointly optimised with a randomly initialised NeRF model.}
\label{tab:supp_refine_llff}
\end{table}

\section{Additional Details}

\subsection{Pseudo Code}
We provide a pseudo code to show the key steps in our pipeline. In addition to set \textit{require\_grad} to \textit{True}, the online ray construction step is another key that enables the direct joint optimisation via back-propagation.

\begin{algorithm}
    \SetAlgoLined
    \SetCommentSty{pseudofont}
    \SetKwFunction{range}{range}
    
    \KwIn{$N$ images $\mathcal{I} = \{I\}_{i=1}^N$ }
    \KwOut{NeRF model $F_\Theta$, camera parameters $[\hat{\bm{\phi}}_i]$, $[\hat{\mathbf{t}}_i]$, and $\hat{f}$}
    
    \BlankLine
    import torch.nn as nn
    
    \BlankLine
    \tcp{Initialisation}
    $[\hat{\bm{\phi}}_i]$ = nn.Parameter(shape=(N, 3), require\_grad=True)\\
    
    $[\hat{\mathbf{t}}_i]$ = nn.Parameter(shape=(N, 3), require\_grad=True)\\
    
    $\hat{f}$ = nn.Parameter(shape=(1,), require\_grad=True)\\
    
    \tcp{NeRF structure see our supp.}
    $F_\Theta$ = NeRF\_Module(require\_grad=True)  
    
    \BlankLine
    \tcp{Training}
    \For{i in \range(N)} 
    {
        \For{m in \range(M)} 
        {
            $\hat{\mathbf{d}}_{i,m}$ = construct\_ray($\hat{\bm{\phi}}_i$, $\hat{\mathbf{t}}_i$, $\hat{f}$, $\mathbf{p}_{i,m}$)  \tcp{Eq. 4}
            
            \BlankLine
            \For{$h$ from $h_n$ to $h_f$}
            {
                $\mathbf{x}_j$ = sample\_point($\hat{\mathbf{d}}_{i,m}$, $\hat{\mathbf{t}}_i$, $h$)
                
                $\mathbf{c}_h$, $\sigma_j$ = $F_\Theta$($\mathbf{x}_h$, $\hat{\mathbf{d}}_{i,m}$)  \tcp{forward NeRF}
                
            }
            $\hat{I}_{i,m}$ = render\_ray($[\mathbf{c}_h]$, $[\sigma_h]$)\\
        }
        
        L = loss($\hat{I}_i$, $I_i$)  \tcp{Eq. 3}
        L.backward()\\
        optimiser.update($[\hat{\bm{\phi}}_i]$, $[\hat{\mathbf{t}}_i]$, $\hat{f}$, $\hat{\Theta}$)
    }
    \caption{\ourmodel Pipeline}
    \label{alg:pipeline}
\end{algorithm}
\vspace{-0.5cm}

\subsection{NeRF Architecture}
\begin{figure}[h]
    \centering
    \includegraphics[width=0.5\textwidth]{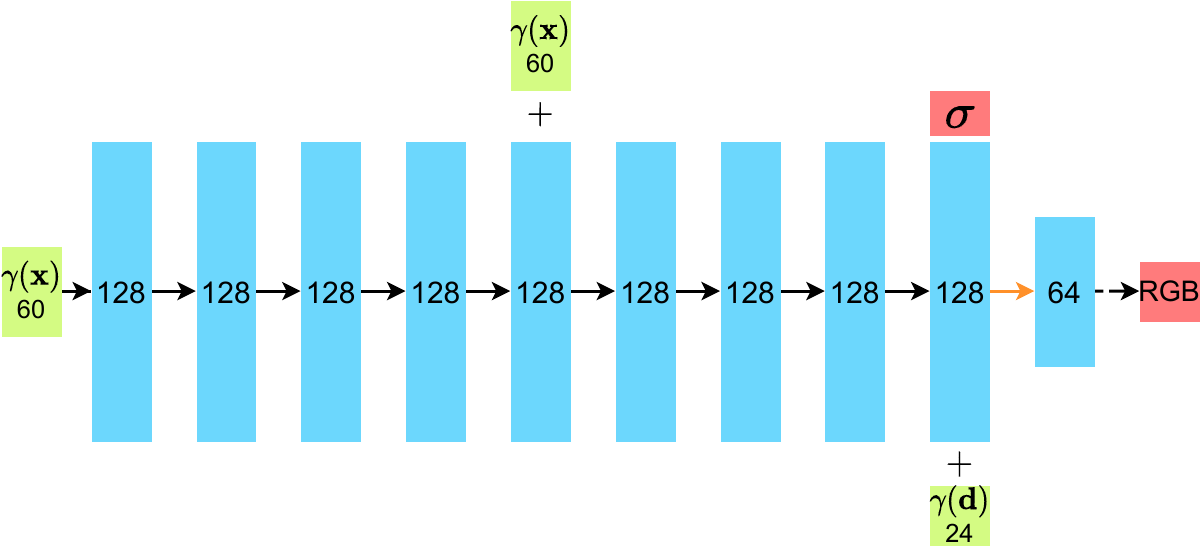}
    \caption{The NeRF implementation used in our paper. Green: position-encoded input. Blue: MLP hidden layers, with channel dimension shown inside. Red: radiance and density outputs. $+$ denotes feature concatenation. Solid black arrow denotes layers with ReLU activation. Dashed black arrow denotes layers with Sigmoid activation. Orange arrow denotes layers without activation. We shrink all hidden layer dimensions in the original implementation by half, i.e. from 256 to 128 for the first 9 layers, and from 128 to 64 for the last layer. We employ the same number of positional encoding frequencies and the same skip links as in original NeRF. The figure style is borrowed from the original NeRF paper so readers can easily make comparisons.}
    \label{fig:our_nerf_arch}
\end{figure}

We employ a smaller NeRF \cite{mildenhall2020nerf} network than the original NeRF paper proposed without a hierarchical structure. Specifically, our NeRF implementation shrinks all hidden layer dimensions by half and follows the same positional encoding and skip connections as implemented in the original NeRF. The network architecture is presented in \cref{fig:our_nerf_arch}.

\subsection{Training Details}
Apart from the implementation details we mentioned in Section 6.1, we report more training details here. Specifically, a BLEFF scene with 27 training images can be trained in 5.5 hours on a single 1080Ti GPU.
This training time is slightly longer than the training time (5 hours) of the original NeRF with our simplifications. The extra half hour training time mainly comes from the online ray direction construction from the learned camera poses and focal length.

\subsection{Evaluation Details} 
Since our optimised camera parameters might lie in different spaces from the ones estimated using COLMAP, for evaluation, we first align the two trajectories globally with a Sim(3) transformation using Umeyama algorithm\cite{umeyama1991least} in an ATE toolbox \cite{Zhang18ATE_ethz}, followed by a more fine-grained gradient-driven camera pose alignment by minimising the photometric error on the synthesised image, while keeping the NeRF model \emph{fixed}. Finally, we compute the metrics between the test image and our synthesised image rendered from the best possible viewpoint. Simply put, all the above mentioned processing aims to eliminate the effect from camera mis-alignment and make a fair comparison on quality of the 3D scene representation.

\subsection{Focal Length Parameterisation}
As mentioned in Section 6.1, we initialise our focal length $f$ to be input image width $W$. In this section, we discuss the focal length parameterisation details, and compare three types of parameterisations. 

In short, we choose to parameterise focal length as a single focal length $f$ with a so-called 2nd-order trick. The details are elaborated in the following subsections.

\subsubsection{$f$ vs. $f_xf_y$}
We choose to optimise a focal length $f$ instead of a horizontal $f_x$ and vertical $f_y$, as we found no performance difference between these choices (\cref{tab:supp_f_vs_fxfy}).

    \begin{table}[h]
    \centering
    \begin{tabular}{llcclcclcc}
    \hline
    \multirow{2}{*}{Scene} &  & \multicolumn{2}{c}{PSNR$\uparrow$} &  & \multicolumn{2}{c}{SSIM$\uparrow$} &  & \multicolumn{2}{c}{LPIPS$\downarrow$} \\ \cline{3-4} \cline{6-7} \cline{9-10} 
     &  & $f$ & $f_xf_y$ &  & $f$ & $f_xf_y$ &  & $f$ & $f_xf_y$ \\ \hline
    Fern &  & 21.67 & 21.79 &  & 0.61 & 0.62 &  & 0.50 & 0.50 \\
    Flower &  & 25.34 & 25.24 &  & 0.71 & 0.71 &  & 0.37 & 0.37 \\
    Fortress &  & 26.20 & 26.24 &  & 0.63 & 0.65 &  & 0.49 & 0.46 \\
    Horns &  & 22.53 & 23.08 &  & 0.61 & 0.63 &  & 0.50 & 0.50 \\
    Leaves &  & 18.88 & 18.79 &  & 0.53 & 0.52 &  & 0.47 & 0.47 \\
    Orchids &  & 16.73 & 16.48 &  & 0.39 & 0.37 &  & 0.55 & 0.56 \\
    Room &  & 25.84 & 25.72 &  & 0.84 & 0.84 &  & 0.44 & 0.44 \\
    Trex &  & 22.67 & 22.53 &  & 0.72 & 0.72 &  & 0.44 & 0.44 \\ \hline
    Mean &  & 22.48 & 22.48 &  & 0.63 & 0.63 &  & 0.47 & 0.47 \\ \hline
    \end{tabular}
    \caption{NVS quality on LLFF-NeRF dataset when using $f$ and $f_xf_y$. We choose to optimise a single focal length $f$ for simplicity, as no performance gain acquired by optimising two focal lengths separately.}
    \label{tab:supp_f_vs_fxfy}
    \end{table}

\subsubsection{1st-order vs. 2nd-order}
A naive way to optimise focal length $f$ is to initialise it with image width $W$ directly, but this poses difficulties to the direct optimisation as $W$ is usually in a much larger numerical magnitude than other learnable parameters.

A better way to parameterise $f$ is through a scale factor $s$:
\begin{equation}
    f = s W,
\end{equation}
and initialise with $s=1.0$. By parameterising focal length with $s$, the network avoids optimising $f$ in pixel unit directly, whose value are often large and pose numerical difficulties in optimisation.

In practice, we found that optimising the square root of $s$, denoted by $\tilde{s}$, leads to slightly better results, \ie
\begin{equation}
    f = \tilde{s}^2 W,
\end{equation}
where $\tilde{s}$ is initialised to 1.0 too. We refer this parameterisation as the 2nd-order trick.
\cref{tab:supp_f_1st_vs_2nd} shows a quantitative comparison of the novel view rendering quality using these two parameterisations.

    \begin{table}[h]
    \centering
    \begin{tabular}{llcclcclcc}
    \hline
    \multirow{2}{*}{Scene} &  & \multicolumn{2}{c}{PSNR$\uparrow$} &  & \multicolumn{2}{c}{SSIM$\uparrow$} &  & \multicolumn{2}{c}{LPIPS$\downarrow$} \\ \cline{3-4} \cline{6-7} \cline{9-10} 
     &  & $\tilde{s}^2$ & $s$ &  & $\tilde{s}^2$ & $s$ &  & $\tilde{s}^2$ & $s$ \\ \hline
    Fern &  & \textbf{21.67} & 21.48 &  & \textbf{0.61} & 0.60 &  & \textbf{0.50} & 0.53 \\
    Flower &  & 25.34 & \textbf{25.44} &  & 0.71 & 0.71 &  & 0.37 & 0.37 \\
    Fortress &  & \textbf{26.20} & 24.92 &  & 0.63 & \textbf{0.58} &  & \textbf{0.49} & 0.58 \\
    Horns &  & 22.53 & 22.53 &  & 0.61 & 0.61 &  & 0.50 & 0.50 \\
    Leaves &  & \textbf{18.88} & 18.85 &  & \textbf{0.53} & 0.52 &  & 0.47 & 0.47 \\
    Orchids &  & \textbf{16.73} & 16.67 &  & 0.39 & 0.39 &  & 0.55 & 0.55 \\
    Room &  & \textbf{25.84} & 25.74 &  & \textbf{0.84} & 0.83 &  & 0.44 & 0.44 \\
    Trex &  & \textbf{22.67} & 22.45 &  & \textbf{0.72} & 0.71 &  & 0.44 & 0.44 \\ \hline
    Mean &  & \textbf{22.48} & 22.26 &  & \textbf{0.63} & 0.62 &  & \textbf{0.47} & 0.49 \\ \hline
    \end{tabular}
    \caption{Quantitative comparison (PSNR on novel views) of two different focal length parameterisations on LLFF-NeRF dataset. The 2nd-order parameterisation $\tilde{s}^2$ produces slightly better results than the 1st-order $s$.}
    \label{tab:supp_f_1st_vs_2nd}
    \end{table}
    
\subsection{RealEstate10K Details}
We use two sequences from RealEstate10K~\cite{zhou2018stereo} in Section 6.2.1 and Section 6.3.2. The details of the these videos and frame rate down-sampling procedures are listed in \cref{tab:supp_re10k_detail}.

\begin{table}[h]
\centering
\begin{tabular}{lllll}
\hline
YouTube video ID & Original fps & Original res. & Training fps & Training res. \\ \hline
MVVJodQ50HQ & 30 & 1920x1080 & 5 & 1920x1080 \\
OT04jHhqYyw & 30 & 1920x1080 & 5 & 1920x1080 \\ \hline
\end{tabular}
\caption{Details for selected sequences in RealEstate10K.}
\label{tab:supp_re10k_detail}
\end{table}

\subsection{BLEFF Asset Licences}
Our BLEFF dataset is made of 14 blender scenes, which we modified and downloaded from \url{https://www.blendswap.com}. The detailed licence info is listed in \cref{tab:supp_bleff_licence}.

\begin{longtable}[c]{llll}
\hline
Scene & Licence & Author & Link \\ \hline
\endfirsthead
\endhead
\hline
\endfoot
\endlastfoot
Airplane & CC-BY-3.0 & fabien & \url{https://blendswap.com/blend/15016} \\
Balls-basketball & CC-BY-3.0 & CGMasters & \url{https://blendswap.com/blend/14758} \\
Balls-volleyball & CC-BY-3.0 & Shri & \url{https://blendswap.com/blend/16958} \\
Bathroom & CC-BY-SA-3.0 & ORBANGeoffrey & \url{https://blendswap.com/blend/20604} \\
Bed & CC-BY-SA-3.0 & Warcos & \url{https://blendswap.com/blend/20975} \\
Castle & CC-BY-0 & BySky & \url{https://blendswap.com/blend/27390} \\
Chair & CC-BY-0 & akashdlfps & \url{https://blendswap.com/blend/25057} \\
Classroom & CC-BY-3.0 & SwastikDas & \url{https://blendswap.com/blend/21410} \\
Deer & CC-BY-0 & Spine69 & \url{https://blendswap.com/blend/26863} \\
Halloween & CC-BY-0 & Yash Deokar & \url{https://blendswap.com/blend/26793} \\
Jugs & CC-BY-0 & BigBadCat & \url{https://blendswap.com/blend/23234} \\
Root & CC-BY-0 & AceTop & \url{https://blendswap.com/blend/24472} \\
Roundtable & CC-BY-0 & gandre82 & \url{https://blendswap.com/blend/26159} \\
Stone & CC-BY-0 & rajatg8008 & \url{https://blendswap.com/blend/22490} \\
Valley & CC-BY-0 & ShadowCrystol & \url{https://blendswap.com/blend/27325} \\ \hline
\caption{BLEFF licence info.}
\label{tab:supp_bleff_licence}\\
\end{longtable}

\section{Broader Impact}
\paragraph{Joint Optimisation Method} Our method focuses on simplifying the classical two-stage NVS systems, by jointly optimising a NeRF model and camera parameters for forward-facing images. We expect our method to be applied in a wide ranged of applications, for example, AR and VR content creation, 3D modelling, camera pose estimation and etc. As this method is designed to be trained on a per-scene basis and is designed for static scenes and cannot handle dynamic objects such as animals or humans, we expect the potential negative social impact (\ie bias in models and human privacy) that can be introduced by our method is very limited, as long as the image capturing procedure confronts local legal requirements.

\paragraph{BLEFF Dataset}
We also introduce a synthetic dataset BLEFF, which contains high-quality path-traced images rendered from 3D Blender models. The BLEFF scenes are selected to cover various common real-life subjects, such as furniture, decorations, toys, rooms, or outdoor landscapes and etc. There are two main purposes for this dataset: 1) to evaluate pose estimation and NVS rendering quality as the same time; and 2) to evaluate the robustness of a joint optimisation system similar to ours. Therefore, we expect our dataset to be beneficial to other joint optimisation systems (NVS and camera parameters) in the future.